\pdfoutput=1

\documentclass[runningheads]{llncs}
\usepackage[utf8]{inputenc}
\usepackage[T1]{fontenc}
\usepackage{hyperref}
\usepackage[all]{hypcap}
\usepackage{times}

\usepackage{multirow}

\hypersetup{
    unicode=true,
	pdftitle={Different Types of Conflicting Knowledge in AmI Environments},
	pdfsubject={Knowledge Representation},
	pdfauthor={Martin Homola, Theodore Patkos},
	pdfkeywords={ambient intelligence, conflicts, classification, resolution}
}

\usepackage[disable]{todonotes}

\newcommand{\assignment}[1]{\todo[color=green!30,inline]{#1}}

\title{Different Types of Conflicting Knowledge in AmI Environments}
\author{%
	Martin Homola\inst{1}
	\and Theodore Patkos\inst{2}
}

\institute{%
Comenius University in Bratislava\\
Mlynsk\'{a} dolina, Bratislava, Slovakia
\and
FORTH-ICS, Heraklion, Greece
}

\begin{document}
\maketitle
\setcounter{page}{37}
\thispagestyle{plain}

\begin{abstract}
We characterize different types of conflicts  that may occur in complex
distributed multi-agent scenarios, such as in Ambient
Intelligence (AmI) environments, and we argue that these conflicts  should
be resolved in a suitable order and with the appropriate strategies for
each individual conflict type. We call for further research with the goal
of turning conflict resolution in AmI environments and similar multi-agent
domains into a more coordinated and agreed upon process.

\assignment{MH: This was now largely reused in the intro and parts are
repeated: possibly cut this down to 3-4 lines}
\end{abstract}

\section{Introduction}
\label{sec:intro}

Ambient Intelligence (AmI) \cite{zelkha:1998,ISTAG03ReducedBib} is a challenging application domains for multi-agent systems, conceived initially to accommodate the ever increasing penetration of interconnected mobile devices into
our everyday life. Ever since, AmI triggered a shift in computing towards developing
more pervasive and sensor-rich environments, often referred to as smart
spaces.  Research in AmI places the human user at the center of
attention aiming at creating intelligent environments with the ability to
adapt to human preferences, serve their needs and goals, and communicate
with their inhabitants utilizing novel means. This paradigm implies a
seamless medium of interaction, advanced networking technology, and
efficient knowledge management, in order to deploy spaces that are
aware of the characteristics of human presence and the diversities of
personalities, being also capable to respond intelligently and
proactively to the users' needs.

AmI environments are populated with embedded computing devices, which can
be abstracted as autonomous agents. These agents are instructed to estimate and
evaluate the current situation and the perceived users' goals, as well as to
cooperate, in order to best support users' needs. Such complex scenarios
require the agents to resolve conflicts that may arise and to act in agreement
\cite{ResendesCS13}. 

In this paper, we look at the knowledge architecture of such systems and analyse the
different types of conflicts that the agents may face: sensory input,
context, domain and background knowledge, goal, and action conflicts. We
argue that such conflicts should be resolved in a suitable
order and using strategies appropriate for each given
type. Moreover, conflicts cannot always be resolved by the agents independently; a certain level of consensus and agreement on conflict resolution
needs to be pursued. This opens new research challenges in knowledge-based
multi-agent systems and their application domains.

\assignment{MH: after finishing Discussion, maybe extend this last
paragraph a little. (If space permits)}

\section{Ambient Intelligence}

Zelka et al. \cite{zelkha:1998} and later Aarts \cite{aarts:2001} devised
the requirements, based on which AmI systems should be:
\begin{enumerate}
\item \emph{embedded} within the environment: users do not need to be concerned
      with their operation,
\item \emph{context-aware:} they are able to recognize the user and the situation,
\item \emph{personalized:} able serve different users according
      to their own needs,
\item \emph{adaptive:} they can change in response to the environment and users
      actions,
\item \emph{anticipatory:} they should understand the users needs and act upon
      them pro-actively and not just in response to user's request.
\end{enumerate}

%
As a multitude of devices serving diverse purposes are
typically installed in smart spaces, it is reasonable to assume that the
agents may be rather heterogeneous in their implementation. Particularly,
their cognitive skills may range from simple reactive agents whose behavior
is based on the most recent sensor readings, to complex knowledge-based and
deliberative agents that perform elaborate reasoning, in order to infer relevant
context, make estimates over the users' intentions, and communicate and
negotiate with the other agents in collaborative manner.
Abstracting away details of implementation, we can generally assume such agents to possess at least the following components:%
\footnote{%
  We take a pragmatic abstraction from the classical BDI agent architecture
  \cite{bdi}: beliefs are stored in the knowledge base, desires
  are mapped as goals, and intentions allow the agent to map the current situation
  into a subset of goals to follow and actions to execute.
}

\begin{itemize} 
\item A knowledge base of some sort, comprising as a
distinguished part the \emph{context model} of the current situation
respective to the agent, and possibly some additional background and domain
knowledge. Each agent may keep track of different aspects of
the world and represent them differently from the other agents.
\item A set of \emph{goals} the agent is able to follow to serve its purpose, from which it selects some subset,
depending on the current perceived context.
\item Either some predefined plans of \emph{actions} to execute 
to achieve each goal or the ability to plan the actions accordingly when
needed.
\item Some means to \emph{communicate} with other agents with the aim to
exchange knowledge and cooperate (e.g., messages, queries, bridge rules,
etc.).
\end{itemize}

It should be remarked that in AmI systems the general aim of an agent is to
perceive and accommodate the goals of the users and to help them in
carrying out actions to achieve these goals. For this reasons, users are likewise often modeled as goal-driven acting agents. This metaphor is indeed useful when studying AmI environments as a
whole, however one must keep in mind that there is a distinction between the
goals of an agent and goals of a user.

\begin{figure}[!t]
\centering
\includegraphics[width=0.5\textwidth]{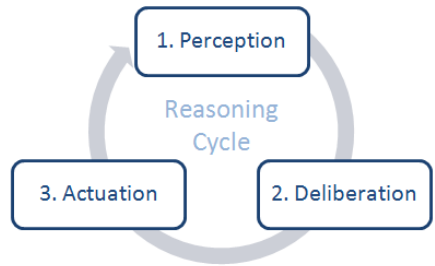}
\caption{Reasoning cycle of autonomous devices in smart spaces}
\label{rcycle}
\end{figure}

An abstract loop that can characterize the basic internal reasoning phases
carried out by an agent is shown in Fig. \ref{rcycle} and involves the
phases of perception, deliberation and actuation. This cycle is triggered
by specific sensory inputs that the agent is monitoring (or the lack of
them) and captures the ability to both deliberate about how best to
\textit{interpret changes} that occur in their dynamically changing world,
as well as to \textit{make decisions} about the most appropriate course of
actions that needs to be taken to support human users' activities. 
%
%
While many approaches have been proposed to study each phase alone, recent
studies (e.g., \cite{PecoraCirilloDUS12,ChenK11}) argue about the need for
a seamless integration of the tasks of perception, recognition and acting
in a coherent loop, in order to synthesize support services in smart
environments with proper and verifiable behavior.

In addition to its dynamic nature, the aspect of heterogeneity is an
equally challenging factor for developing AmI services. Agents operating in
smart spaces may have different reasoning skills, obtain access to distinct
knowledge repositories, local or shared, and evaluate incoming information
based on different trust criteria. A real-world smart system needs to
respect the fact that the way context is inferred by each
involved agent is not an objective process. Being highly distributed, these
environments produce information that can be interpreted in a totally
different manner by the various intelligent agents; and it is not
uncommon for the agents to end up having incoherent and conflicting views
of the current context.

\section{Conflicting Knowledge}
\label{sec:conflict_intro}

The importance of dealing with conflicts has been noted by
other researchers working in AmI
\cite{ResendesCS13,HenricksenMUI04ReducedBib,Ortega10ReducedBib}. Resendes et al.
\cite{ResendesCS13} analyze different types of
conflicts that may arise in AmI systems and organize them into a taxonomy,
as listed in Table~\ref{tab:resendes-conflicts}.

\begin{table}[thp!]
\caption{Taxonomy of conflicts \cite{ResendesCS13}}
\label{tab:resendes-conflicts}
\centering
\begin{tabular}{|l||l|l|l|l|}
\hline
\textbf{Dimension} & Source & Intervenients & Detection time & Solvability \\
\hline
\multirow{4}{*}{\textbf{Possible types}} 
   & resource	 & single user	   & a priori	    & conflict avoidance\\
   & application & user vs.\ user  & when it occurs & conflict resolution\\
   & policy 	 & user vs.\ space & a posteriori   & acknowledge inability\\
   & role	 &		   &		    & acknowledge occurrence\\
\hline
\end{tabular}
\end{table}

The authors identify four basic broad categories of conflicts, which are dubbed
\emph{dimensions}, in order to stress their orthogonality, i.e., the fact
that one conflict can be independently classified with respect to each of them. 
The
\emph{source} dimension indicates where/how each conflict originates -- it
may be the case that users (or applications) are conflicting over some
resource allocation, or it is not possible to execute some action due to
policy, or there are conflicting user profiles. Following the
\emph{intervenients} dimension, there might be conflicting intentions
within a single user, between multiple users, or between user and the
space. The \emph{detection time} dimension sorts conflicts into those
that are (can be) detected a priori, at the time they occur, or only a
posteriori. Finally, the \emph{solvability} dimension indicates at which
level can conflicts be resolved -- before they happen (i.e., to avoid them), immediately when they happen, or after some delay, in which case they are further split into those conflicts which cannot be
resolved at all and those which cannot be resolved due to being detected
too late.

The taxonomy of Resendes et al. is arguably very useful. In addition to this classification, though, and considering the
agent architecture of AmI systems, it appears to us that conflicts should
also be categorized based on the different types of knowledge in which they
appear.  This is due to the fact that each type of knowledge is processed
differently, and in a different point of the agents reasoning cycle
depicted in Fig.\,\ref{rcycle}. This classification
can be seen as yet another dimension, orthogonal to the previously
discussed four, which we propose to add into the taxonomy of Resendes et
al., as shown in Table~\ref{tab:resendes-knowledge-type} and described next:

\begin{table}[thp!]

\caption{The knowledge dimension of conflicts}
\label{tab:resendes-knowledge-type}
\centering
\begin{tabular}{|l||l|}
\hline
\textbf{Dimension} & Knowledge type\\
\hline
\multirow{4}{*}{\textbf{Possible types}}		& sensory input \\
                        		& context \\
                        		& domain/background \\
					& goal \\
					& action \\
\hline
\end{tabular}
\end{table}


\begin{description}
\item[Sensory input conflict:] if a conflicting reading of some sensors
appears. This type may refer to multiple readings of the same or similar sensors or it may be the result of different sensors, whose outputs are mutually exclusive (the agents know that these outputs cannot
occur at the same time). The conflict may arise within a single agent, but
it may also be distributed between more then one agent (each containing
part of the conflicting readings). The latter option may subsequently
cause a contextual conflict.
\item[Contextual conflict:] if two (or more) agents are part of the same
situation, their models of the world may be conflicting,
implying, e.g., a different location, or perceived activity of the user,
etc. This type of conflict is may be caused by a previous unnoticed
sensory input conflict, but also by a different evaluation
of the situation.
\item[Domain and background knowledge conflict:] domain and background
knowledge refer to the information the agent possesses and uses, in order to
fulfill its purpose.
Conflicts in these types of knowledge, if they occur, may require a
different kind of solution. In contrast with contextual information which
is dynamic and changing, domain and background knowledge are often
considered unchanging and fully specified (to the extent required by the
application). Hence, redesign of the agent's knowledge base by its creator
may be required, rather than resolving the conflict in an automatic manner. 
\item[Goal conflict:] if two (or more) agents are part of the same
situation, their models of the world are compatible, but they have
mutually conflicting goals. Note that we do not consider it a goal conflict
if agents have conflicting goals in different models of the world, because
it is natural to have different goals in different situations.
\item[Action conflict:] if two (or more) agents share a compatible model of
the world, and a compatible set of goals, however, they decide to undertake
a conflicting course of actions to carry our their goals.
\end{description}

\assignment{MH: add some closing paragraph/bridge to the next section}

\section{Discussion}
\assignment{MH: make the actual argument here: (a) conflicts need to resolved
in certain order and in appropriate order; (b) conflicts need to be
resolved in agreement, and not independently like some current approaches
do it, current KR approaches need to be extended, to allow this... etc}

The ability to efficiently deal with conflicts is imperative, in order to
appropriately balance between the two main design principles that have been
set for the success of AmI systems: being as less intrusive as possible with minimal need for human intervention, while still allowing users to feel
confident that they are in control.  Devising intelligent automated
mechanisms for identifying, preventing or resolving conflicts is of utmost
importance in this area of research.

Compared with the classic BDI architecture, the proposed knowledge type
dimension is more fine gained: the agent's representation of sensory
inputs, context, and domain/background knowledge are all different types of
beliefs. As we argued, it is important to distinguish between them because
they are resolved differently, and in different time.
For instance, together with the two additional knowledge types they can be
sorted on the scale from lower to higher level of knowledge: (a) sensory
input, (b) contextual, domain and background knowledge, (c) goals, and (d)
actions, in the respective order. Distinguishing between these five types
is important also due to the following conjecture: solving conflicts in a
lower level knowledge can reduce and may possibly prevent occurrence of further conflicts
in the higher levels of knowledge. For example, if two agents
have a conflict in the contextual knowledge, that is, their interpretation
of the situation in which they both participate is not compatible (e.g.,
they may have conflicting information about location) -- if the conflict is
resolved at this level, it is less likely that the agents will come up with
conflicting goals and consequently action plans.

Due to different nature and complexity of conflicts at the different
knowledge types, different formalisms and tools are suitable to resolve
each type of conflict in this respect. Sensory conflicts are
probably most effectively resolved with use of repeated and redundant
readings which are then cleaned using statistical methods \cite{LuF09}.
Hybrid approaches that combine statistical methods with reasoning were
applied on resolving contextual conflicts (e.g., for identifying the user's
situation \cite{YeDMcK12}).  As we noted above, conflicts in domain and
background knowledge most likely require a manual solution.

As noted above, a number of higher level conflicts can likely be prevented
by timely identification and resolution of conflicts at lower levels. The
remaining goal and action conflicts are more intriguing. While approaches
such as multi-context systems \cite{giunchiglia:1993,brewka:2007b} allow to
build agents capable to deliberate on knowledge obtained from other agents
\cite{sabater:mcs-agents,BikakisA10TKDE}, the conflict resolution process
is typically confined within an agent, and executed independently from the
other agents. Two independent agents may thus resolve the same conflict
differently without a deeper consensus, resulting into ill-coordinated
action. This points out that further investigations are needed in order to
develop multi-agent architectures with consensual and cooperate conflict
resolution. Few existing works
\cite{FerrandoDAM12reducedBib,BikakisA10TKDE,Moraitis2007,brewka:arg-mcsReducedBib} show a
prospective path by incorporating results from areas such agreement
technologies \cite{agreement}, argumentation \cite{Dung}, computational
social choice \cite{comp-soc-choice}, etc.

\paragraph{Acknowledgements.}
This work resulted from the Slovak--Greek bilateral project ``Multi-context
Reasoning in Heterogeneous environments'', registered on the Slovak side
under no. SK-GR-0070-11 with the APVV agency and co-financed by the Greek
General Secretariat of Science and Technology and the European Union.  It
was further supported from the Slovak national VEGA project no.~1/1333/12.
Martin Baláž and Martin Homola are also supported from APVV project
no.~APVV-0513-10.

\assignment{MH: $\uparrow$ missing projects}

\fontsize{9.5pt}{10.5pt} \selectfont
\bibliographystyle{splncs03}
\bibliography{biblio}

\end{document}